\documentclass[journal]{IEEEtran}

\ifCLASSINFOpdf
\else
\fi
\usepackage{etoolbox}
\makeatletter
\patchcmd{\@makecaption}
{\scshape}
{}
{}
{}
\makeatother
\usepackage{algorithm,algpseudocode,float}
\usepackage{lipsum}

\usepackage{algorithm}
\usepackage{algpseudocode}
\usepackage{amsmath}
\usepackage{lineno,hyperref,multirow,graphicx,subfigure,enumerate,color,amsmath, epstopdf}
\usepackage{booktabs}
\usepackage[justification=centering]{caption}
\usepackage[noadjust]{cite}
\usepackage{stfloats}
\usepackage[utf8]{inputenc}
\usepackage[greek,russian,spanish,english]{babel}
\usepackage[OT2,T1, T2A,OT1]{fontenc}
\usepackage{stfloats}
\begin{document}
	
	\title{GBC: An Efficient and Adaptive Clustering Algorithm Based on Granular-Ball}
	
	\author{Shuyin~Xia,
		Jiang~Xie$^{*}$,
		Guoyin~Wang
		
		\thanks{Shuyin Xia, Jiang Xie, Guoyin Wang are with the Chongqing Key Laboratory of Computational Intelligence, Chongqing University of Telecommunications and Posts, 400065, Chongqing, China. E-mail: xiasy@cqupt.edu.cn, xiejiang@cqupt.edu.cn (corresponding author), wanggy@cqupt.edu.cn.}}
	
	\maketitle
	
	\begin{abstract}
		Existing clustering methods are based on a single granularity of information, such as the distance and density of each data. This most fine-grained based approach is usually inefficient and susceptible to noise. Inspired by adaptive process of granular-ball division and differentiation, we present a novel clustering approach that retains the speed and efficiency of K-means clustering while out-performing time-tested density clustering approaches widely used in industry today. Our simple, robust, adaptive granular-ball clustering method can efficiently recognize clusters with unknown and complex shapes without the use of extra parameters. Moreover, the proposed method provides an efficient, adaptive way to depict the world, and will promote the research and development of adaptive and efficient AI technologies, especially density computing models, and improve the efficiency of many existing clustering methods.
	\end{abstract}
	
	
	\IEEEpeerreviewmaketitle
	
	\section{Introduction}
	
	\IEEEPARstart{C}{lustering} is a popular approach for understanding complex unlabeled data, which are commonly produced by empirical studies and are not easy to analyze manually\cite{1,2}. Partition-based clustering methods\cite{3} assume that clusters are exclusive groups of data characterized by small, within-cluster sums of distances, and the similarity in such groups should be maximized. K-Means is the representative algorithm of partition-based clustering methods, and classical K-Means clustering achieves good clustering results on datasets with convex spherical structures\cite{4}. Methods based on density can overcome these limitations of the assumption. Density-based methods reconstruct arbitrary shapes according to a density-connectivity criterion. The most widely used density-based method is DBSCAN (density-based spatial clustering of applications with noise)\cite{5}, which considers two points to belong to the same cluster if enough points in a neighborhood are common (density reachable). However, when two or more clusters are present and in proximity, a wide density reachability threshold may join them, while an excessively strict threshold may fail to detect clusters\cite{6}. Recently, Rodriguez and Laio proposed a popular method that achieves clustering by finding density peaks (DPeak)\cite{7}. DPeak addresses the limitations of DBSCAN by initially finding density peaks and using them to separate clusters. This method assumes that cluster centers are surrounded by neighbors with lower local density and that they are at a relatively large distance from any point with higher density. DPeak can recognize clusters regardless of their shape and the dimensionality of the space in which they are embedded, but it lacks an efficient quantitative criterion for judging the cluster center.
	
	Since what we propose in this paper is a new and basic algorithm, and K-means, DBSCAN and DPeak are three most popular original clustering algorithms, so as that in\cite{7}, only these algorithms are used for comparison in this paper. Both DBSCAN and DPeak, along with their peers, face two challenges. First, they are not adaptive to clusters with different densities. Second, the performance of DBSCAN and DPeak is sensitive to hyperparameters, and it is nontrivial to correctly set these parameters for different datasets\cite{8}.
	
	The main contributions of this paper are as follows:
	
	\hangafter=1
	\setlength{\hangindent}{2em}
	1) Self-adaption:  The generation of coarse-grained granular-balls is based on the data distributed measurement, which is completed through adaptive iteration.
	
	\hangafter=1
	\setlength{\hangindent}{2em}
	2) Efficiency: Because the number of granular-balls is far less than the number of data, the clustering algorithm based on granular-ball has a high performance improvement.
	
	\hangafter=1
	\setlength{\hangindent}{2em}
	3) Robustness: Since each granular-ball covers many points and only contains two data, namely center and radius, a small number of noise points can be smoothed, so that the granular-ball will not be affected.
	
	\section{Related Work about Granular-ball Computing}
	In response to the related challenges of clustering mentioned in Section 1, we propose granular-ball clustering: an adaptive clustering method based on the natural process of granular-ball division and differentiation\cite{9,10} and partly inspired by the "large scale first" cognitive mechanism\cite{11}. It is different from the major existing artificial intelligence algorithms, which take the most fine-grained points as input. The human brain's global precedence cognition is efficient and robust, and is very beneficial for improving the performance of the existing artificial intelligence algorithms. The generation model of granular-ball is as follows.
	
	Given a data set $D = {p_i(i=1, 2, ..., n)}$, where $n$ is the number of samples on $D$. Granular balls $GB_1, GB_2, \dots , GB_m$ are used to cover and represent the data set $D$. Suppose the number of samples in the $j^{th}$ granular-ball $GB_j$ is expressed as $|GB_j|$, then its coverage degree can be expressed as $ {\textstyle \sum_{j=1}^{m}}\left ( \left | GB_j \right |  \right ) /n $. The basic model of granular-ball coverage can be expressed as
	\begin{equation}\label{eqGB}
		\setlength{\abovedisplayskip}{6pt}
		\setlength{\belowdisplayskip}{3pt}
		\begin{aligned}
			min \ \ \lambda _1 \ast n /{\sum_{j=1}^{m}}\left ( \left | GB_j \right |  \right ) /n + \lambda _2 \ast m,\\ 
			s.t. \ \ quality(GB_j) \ge T, \ \ \ \ \ \ \ \ \ \ \ \ 
		\end{aligned}
	\end{equation}
	where $\lambda _1$ and $\lambda _2$ are the corresponding weight coefficients, and $m$ is the number of granular balls. When other factors remain unchanged, the higher the coverage, the less the sample information is lost, and the more the number of granular-balls, the the characterization is more accurate. Therefore, the minimum number of granular-balls should be considered to obtain the maximum coverage degree when generating granular-balls. By adjusting the parameters $\lambda _1$ and $\lambda _2$, the optimal granular-ball generation results can be obtained to minimize the value of the whole equation. In most cases, the two items in the objective function do not affect each other and do not need trade off, so $\lambda _1$ and $\lambda _2$ are set to 1 by default. Granular-ball computing can fit arbitrarily distributed data.

	The process of granular-ball division can produce an appropriate number of granular-balls. Based on these granular-balls, clusters can be formed through granular-ball differentiation which merges adjacent granular-balls. The process we model after granular-ball division produces a number of granular-balls, which are represented using what we call granular-balls. These granular-balls differentiate into different clusters by merging adjacent granular-balls in our algorithm. The clustering process consisting of granular-ball division and differentiation is shown in Figs. 1(A-F). The data set in Fig. 1 can be seen as two clusters. In Fig. 1(A), the data set is observed as a single granular-ball. In Figs. 1(B-C), as the granular-balls divide, the clusters grow and form. In Fig. 1(D), two clusters have taken shape and matured. In Fig. 1(E), granular-balls that are close together form an cluster in Fig. 1(F).
	
	\begin{figure}[H]
		\vspace{0em}
		\includegraphics[width=0.48\textwidth]{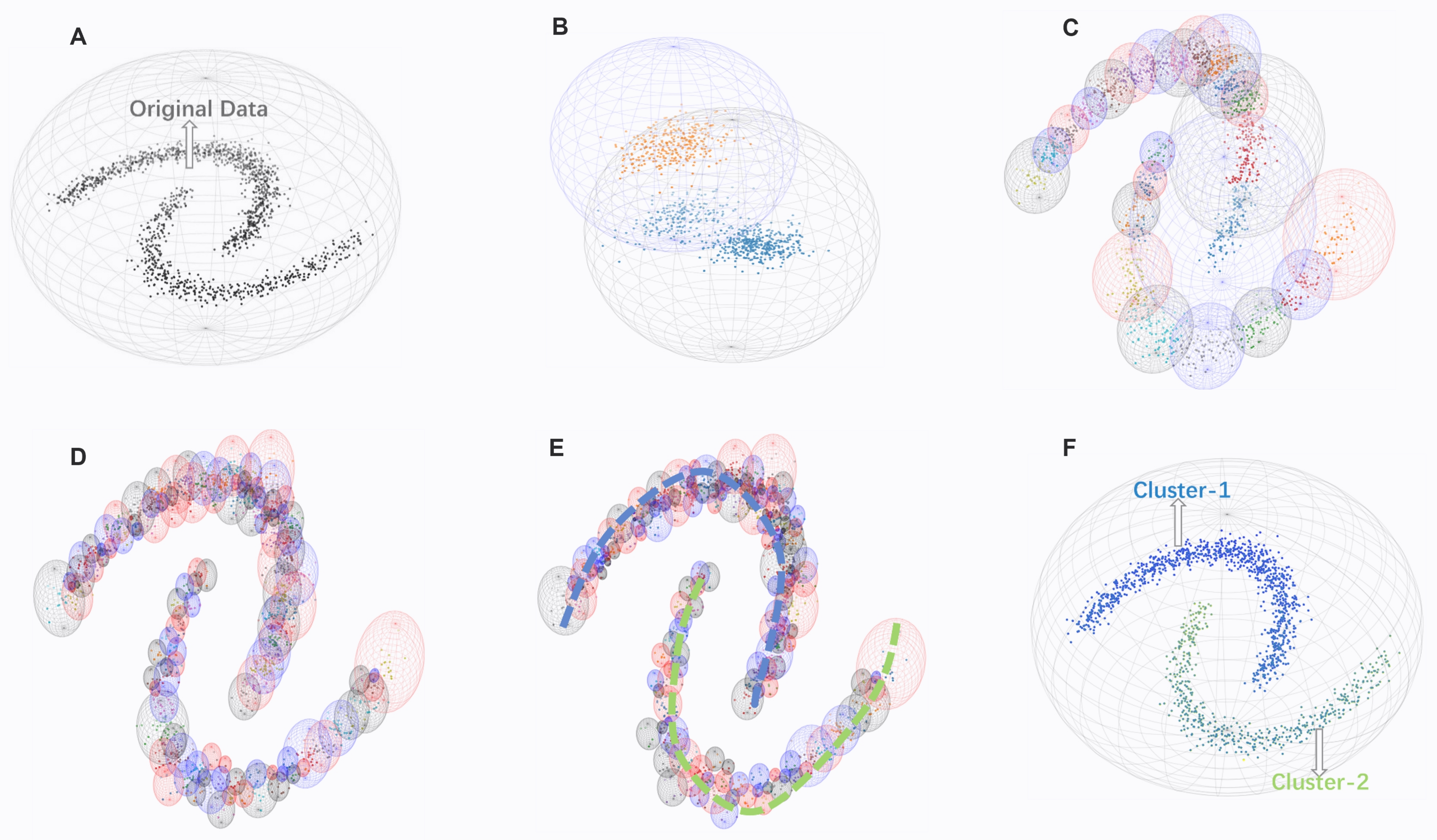}
		\captionsetup{name={Fig.},labelsep=period,singlelinecheck=off,font={small},justification=raggedright}
		\caption{The process of granular-ball clustering.  (A) Point distribution for the Moon dataset. (B)-(D) The granular-ball division process. (E) The granular-ball differentiation process. (F)  The clustering results.}
		\label{fig:1}
		\vspace{0em}
	\end{figure}
	
	\section{Granular-Ball Clustering}
	
	The previous work\cite{9,10,12,13} on granular-ball computing is mainly for supervised learning. For the clustering method, we combine formula (1) and heuristic rules to build the quality evaluation and splitting rules for granular-ball.
	A granular-ball is used to represent a granular-ball. Its definition is as follows. Let $D \in R^d$ be a dataset and $P_i =(p_{i1},p_{i2},\dots,p_{id})\in D$ be a datapoint, where $p_{il}$ denotes the coordinate value of $P_i$ on the $l$-th dimension. A granular-ball ($GB_k$) is defined as:${GB}_{k} = \left\{ P_{i} \middle| \right\|\left. P_{i} - c_{k} \right\|\left. \leq r_{k} \right\}$, where $ c_k $ is the center of gravity of all data points in $ GB_k $, and $ r_k $ is the maximum distance from all points $ P_i $ in  $ GB_k $ to $  c_k $, i.e., that,
	\begin{equation}
		\label{equ:1}
		{c_{k} = \frac{1}{n_{k}}{\sum_{i = 1}^{n_{k}}P_{i}}~,~r_{k} = max \|P_{i} - c_{k}\|,~i = 1,2\ldots,n_{k},}
	\end{equation}
	where $\left\| . \right\|$ denotes the L2-norm throughout the paper, and $n_k$ is the number of data points in $ GB_k $. $ c_k $ can be seen as the nucleus of the granular-ball $ GB_k $, and $ r_k $ the size of the granular-ball.
	
	After the granular-ball is defined, we need to measure the quality of a granular-ball to know when the granular-ball should stop dividing. For any granular-ball $ CB_k $, the quality is measured by the average distance $AD_k$. The average distance is calculated by computing the ratio of the sum radius $s_{k} = {\sum_{i = 1}^{n_{k}}\left\| P_{i} - c_{k} \right\|}$ and the number of datapoints $ n_k $ in $ CB_k $, that is:
	
	\begin{equation}
		\label{equ:2}
		{{AD}_{k} = \frac{s_{k}}{n_{k}},}
	\end{equation}
	
	\begin{figure}
		\centering
		\includegraphics[width=0.50\textwidth]{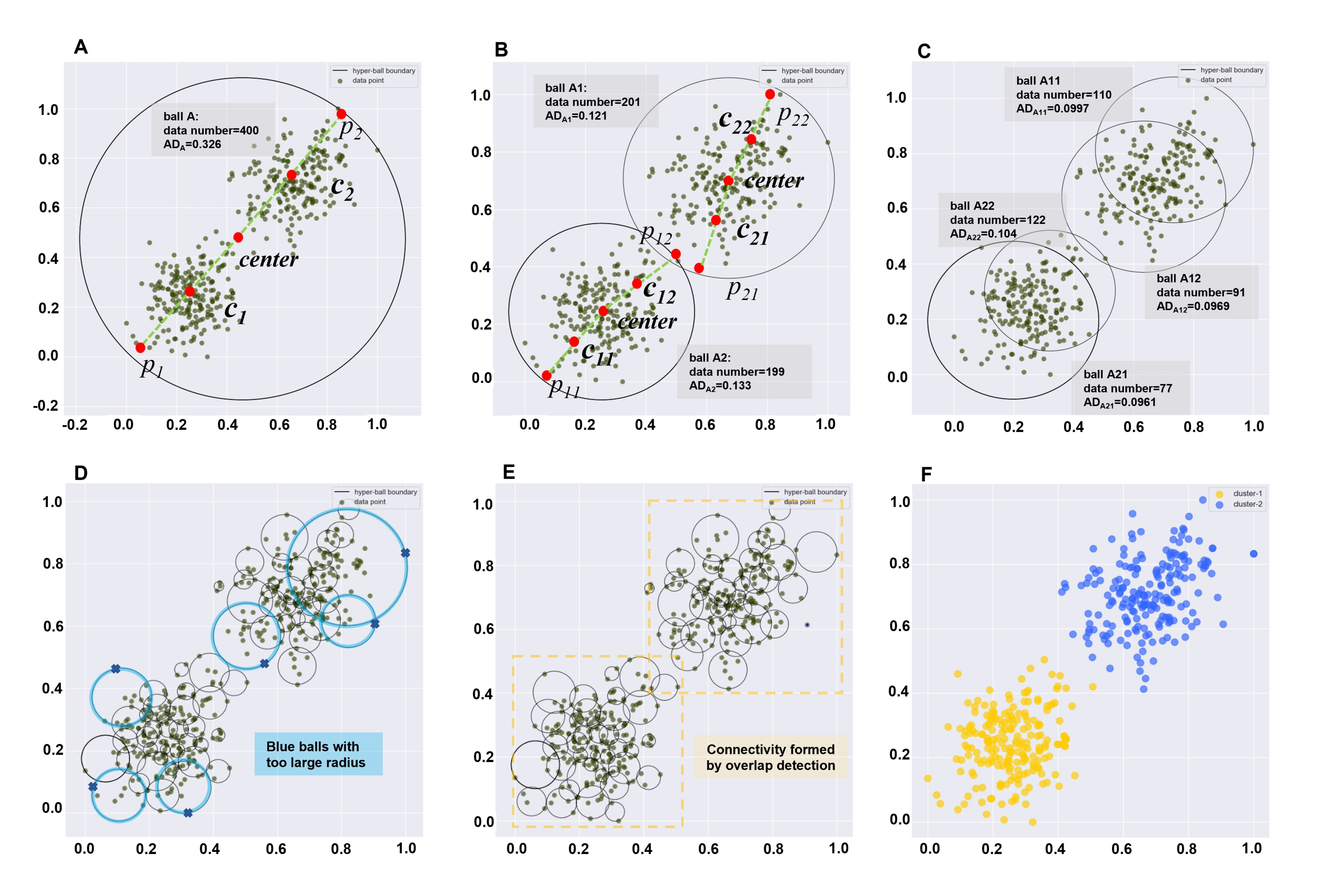}
		\captionsetup{name={Fig.},labelsep=period,singlelinecheck=off,font={small},justification=raggedright}
		\caption{The Division and differentiation Process on a simplified example. (A) Raw data points in a granular-ball $A$ ($GB_A$) with 400 points and $AD_A$ is 0.326. (B) Child granular-balls of $A$. (C) Child granular-balls of $A_1$ and $A_2$. (D) Blue balls with too large radii. (E) The final granular-ball sets after removing granular-balls with too large radii. (F) Granular balls are differentiated into two different clusters.}
		\vspace{-1.0em}
		\label{fig:2}	
	\end{figure}
	
	$AD_k$ describes the average proximity (represented by the average distance) of all points in a granular-ball to the center $c_k $. The smaller the average distance $AD_k$, the more similar the points inside the granular-ball, and the higher the granular-ball quality. The point at which a granular-ball stops dividing is determined by the following adaptive rule: if the average distance of a granular-ball is smaller than that of each granular-ball it would divide into, it has reached a high enough quality and stops dividing. The entire division and differentiation process is illustrated in an example in Fig. 2. In Fig. 2(A), we have granular-ball A ($GB_A$) with an average distance value $AD_A$ of 0.326. In granular-ball $GB_A$, $p1$ is the farthest point from the center, and $p2$ is the farthest point from $p1$, here we select the mid-point $c1$, $c2$ of $p1$, $p2$ from center as the initial cluster center. As shown in Fig. 2(B), $GB_A$ can be split into $GB_{A_1}$ and $GB_{A_2}$ by executing one division according to the distances from all points to $ c1 $ and $ c2 $. Since $AD_{A_1}$ and $AD_{A_2}$ are smaller than the average distance of $GB_A$, the division operation will proceed. Ball division continues until no granular-balls split: in Fig. 2(D), the final division result is illustrated. However, in Fig. 2(D), some granular-balls with overlarge radii are malformed and intermediate, and have not yet divided into a normal mode. These types of granular-balls are usually generated by boundary or noise points, and will split continually. A granular-ball of this type can be characterized by a radius:
	\begin{equation}
		\label{equ:3}
		{{r_{k} > 2*max\left( mean(r),median(r) \right)},}
	\end{equation}
	where mean(r) and median(r) represent the mean and median of all granular-ball radii, respectively. After granular-balls with excessively large radii are removed through further splitting, the division process is completed, and the result is shown in Fig. 2(E).
	
	When each granular-ball has a high enough quality and stops dividing, adjacent granular-balls must merge clusters. The adjacency rules are described as follows. Let $ o_i $ represent the cumulative value of the overlapping times between $CB_i$ and its neighbor granular-balls, and $ o_j $ represent the cumulative value of the overlapping times between $CB_j$  and its neighbor granular-balls. Let $\tau _{ij}$ be an adjustment coefficient given by
	
	\begin{equation}
		\label{equ:4}
		{{\tau_{ij} = \frac{\left( {r_{i},~~~r}_{j} \right)~}{1 + min\left( {o_{i,}{~o}_{j}} \right)}},}
	\end{equation}
	and does not need to be optimized. Then the adjacency criterion for two granular-balls is given by
	\begin{equation}
		\label{equ:5}
		{{\left\| c_{i} - c_{j} \right\| - \left( {r_{i} + r_{j}} \right) < \tau_{ij}},}
	\end{equation}
	
	The formula shows that if the gap between $GB_i$ and $GB_j$ is less than  $\tau _{ij}$, they meet the adjacency criterion. $\tau _{ij}$ can dynamically adjust the criterion during the algorithm. If two granular-balls satisfy the adjacency criterion, they belong to the same cluster. $\tau _{ij}$ reflects the situation that the more times a granular-ball overlaps with its neighbor granular-balls, the higher the probability that the granular-ball and its neighbor granular-balls form a single cluster, and the lower the probability that the granular-ball and other granular-balls that do not overlap with it become a cluster.
	
	We demonstrate this process of organ formation using the adjacency criterion and the adjustment coefficient in Fig. 3. In Fig. 3(A),  $GB_2$ overlaps its two nearest neighbor granular-balls; in Fig. 3(B),  $GB_2$ has no overlapping nearest neighbor granular-balls. Therefore, in Fig. 3(A),  $GB_2$ has more neighbor granular-balls than that in Fig. 3(B), and is likely to generate an independent organ cluster; in Fig. 3(B), $GB_2$ is more likely to be absorbed into other clusters (i.e., $GB_1$). Accordingly, the  $\tau _{12}$ in (A), 0.166, is smaller than the $\tau _{12}$   in (B), 0.5. When using the definition of average distance $AD_k$, the division of granular-balls with a large radius may produce some very small granular-balls, noise granular-balls, containing only one data point with a radius close to zero. These noise granular-balls are considered to be too small and have no influence, so they do not participate in the differentiation process. Instead, after the differentiation process is completed, they are assigned to the nearest cluster or marked as noise.
	\begin{figure}[t]
		\centering
		\includegraphics[width=0.48\textwidth]{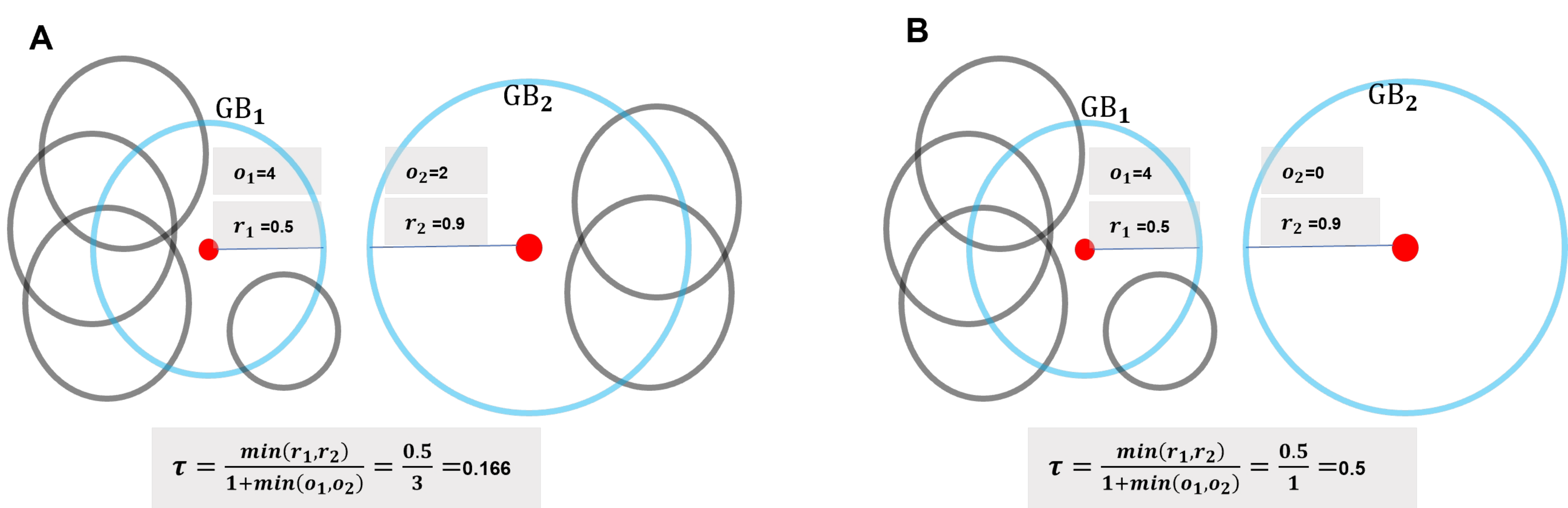}
		\captionsetup{name={Fig.},labelsep=period,singlelinecheck=off,font={small},justification=raggedright}
		\caption{ $o_1$ and $o_2$ are the cumulative values of the overlapping times of the neighbor granular-ball as the adjustment coefficients of $\tau$, which can dynamically adjust the criterion that meets the adjacent rules. To prevent the wrong differentiation of the granular-balls, the larger the cumulative value is, the stricter the standard that meets the adjacent rules. The $\tau$ in $A$ is smaller than the  in $B$.}
		\vspace{-1.0em}
		\label{fig:31}	
	\end{figure}
	
	To sum up, the process of granular-ball division from large to small is conducive to finding noise points, rendering the method robust to noise. The final clustering result based on Fig. 2(E) is illustrated in Fig. 2(F). Compared with other clustering methods, the clustering process does not contain any hyper-parameters or optimize any parameters, fully adaptive. In addition, compared with most other clustering methods that need repeated iterative optimization, the final merging process of clustering is completed in one step. Furthermore, in comparison with most clustering methods based on density, such as DBCAN and DPeak, GBC does not need to calculate the distance between each pair of two data objects. So, GBC is very efficient.

	\section{Experiment}
	
	\begin{figure}
		\centering
		\subfigbottomskip=-2pt
		\subfigure{
			\includegraphics[width=0.4\textwidth]{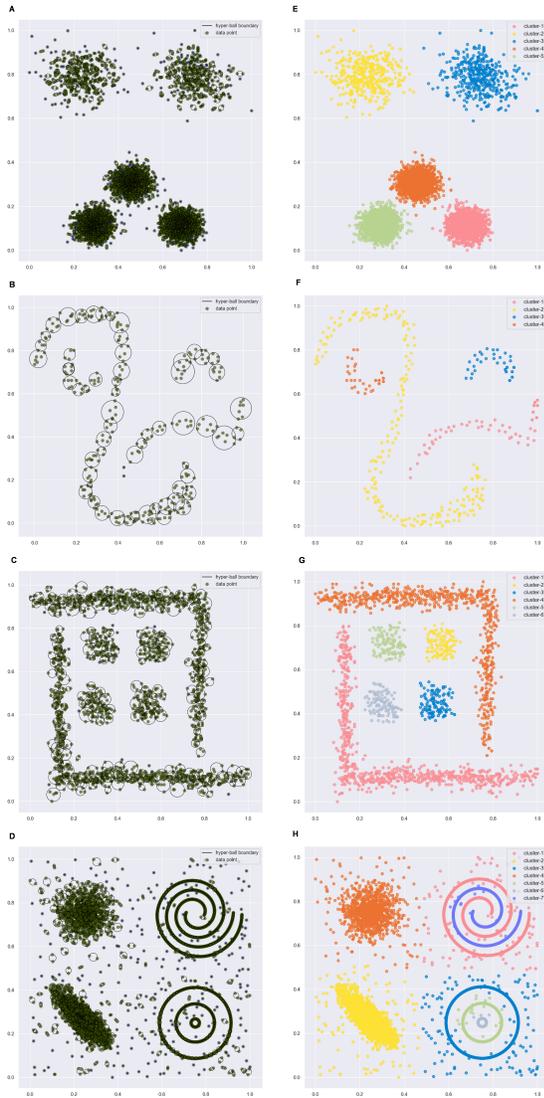}}
		\captionsetup{name={Fig.},labelsep=period,singlelinecheck=off,font={small},justification=raggedright}
		\caption{Evaluation on synthetic data in the literature (6).  (A)-(D) Granular-ball sets for raw data points. (E)-(H) Clustering results of GBC.}
		\vspace{-1.0em}
		\label{fig:nan}	
	\end{figure}
	We quantitatively evaluated our method on several common synthetic datasets from clustering literature\cite{14} . Figs. 4(A-D) show the final granular-ball refinement results of the datasets. In Fig. 4(A), we consider five clusters with large variations in density. The densities of the two clusters in the top half are sparser, and the bottom three clusters are denser. Our method correctly determines the cluster structure of the dataset. In Fig. 5(B) and Fig. 4(C), the clustering results demonstrate that our method can process manifold datasets. The dataset in Fig. 5(D) is composed of three inscribed circle clusters, two spiral tangled clusters, and two spherical clusters with a total of 8533 objects, including noise objects. The clustering result in Fig. 4(D) shows that our method can obtain results comparable to those of the original article and is robust to noise. As a comparison, we show the cluster results obtained by K-Means, DBSCAN, and DPeak for these four synthetic datasets in Fig. 5, Fig. 6, and Fig. 7. Even when K-Means, DBSCAN, and DPeak are optimized repeatedly with optimized parameter settings, in many cases the clustering results are not compliant with visual intuition; in contrast, although GBC does not need to optimize any parameters, the clustering results are significantly better, and obviously visually intuitive.
	\begin{figure}[H]
		\centering
		\subfigbottomskip=-2pt
		\subfigure{
			\includegraphics[width=0.3\textwidth]{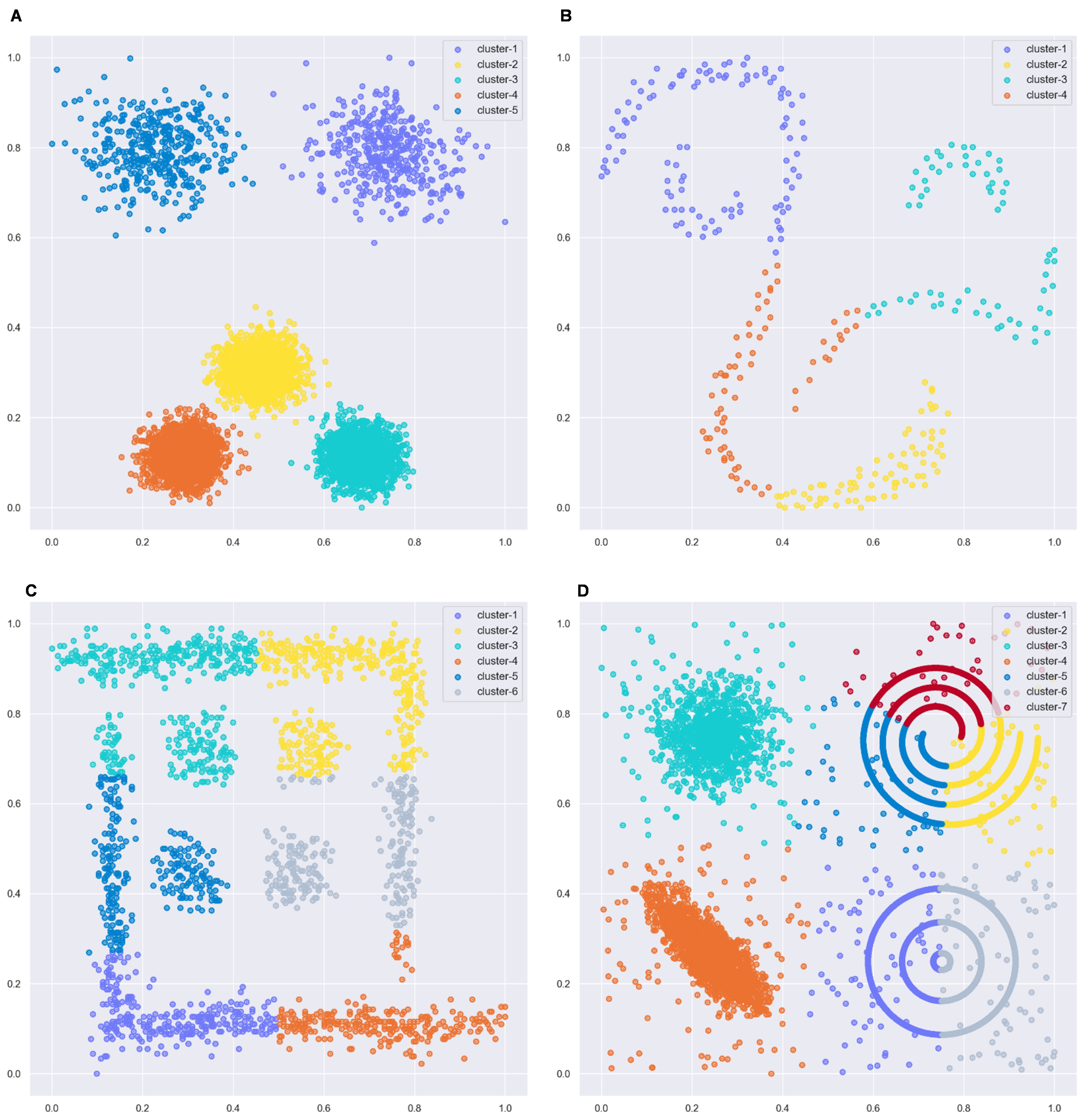}}
		\captionsetup{name={Fig.},labelsep=period,singlelinecheck=off,font={small},justification=raggedright}
		\caption{The clustering results of K-Means on 4 synthetic datasets. In all the cases, the value of K has been chosen by visual inspection.}
		\vspace{-1.0em}
		\label{fig:nan}	
	\end{figure}
	
	\begin{figure}[H]
		\centering
		\subfigbottomskip=-2pt
		\subfigure{
			\includegraphics[width=0.3\textwidth]{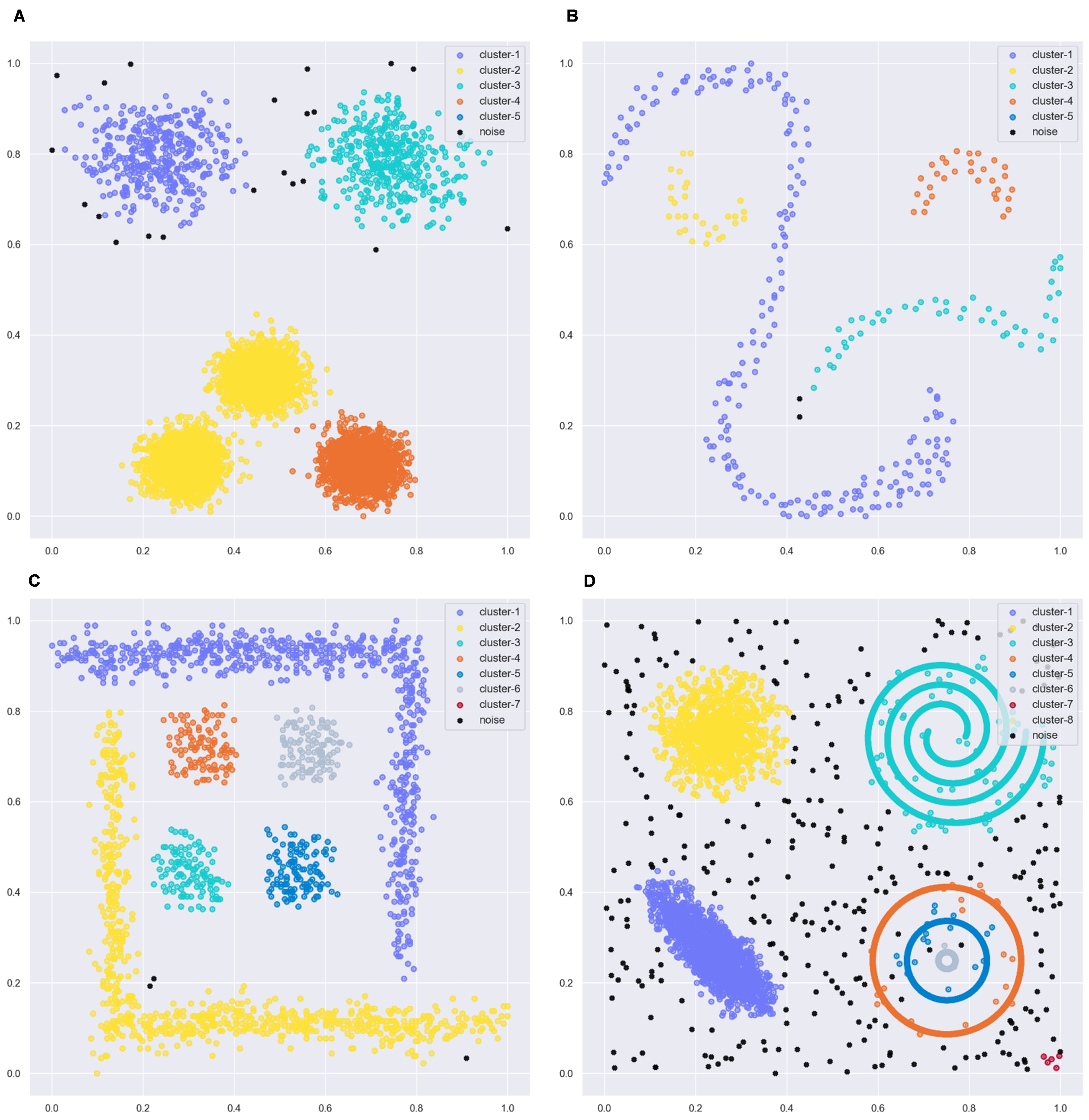}}
		\captionsetup{name={Fig.},labelsep=period,singlelinecheck=off,font={small},justification=raggedright}
		\caption{The clustering results of DBSCAN on 4 synthetic datasets. The clustering results have been obtained by running 1000 times the algorithm and taking the best solution according to the different eps and minpts.}
		\vspace{-1.0em}
		\label{fig:nan}	
	\end{figure}
	\begin{figure}[H]
		\centering
		\subfigbottomskip=-2pt
		\subfigure{
			\includegraphics[width=0.3\textwidth]{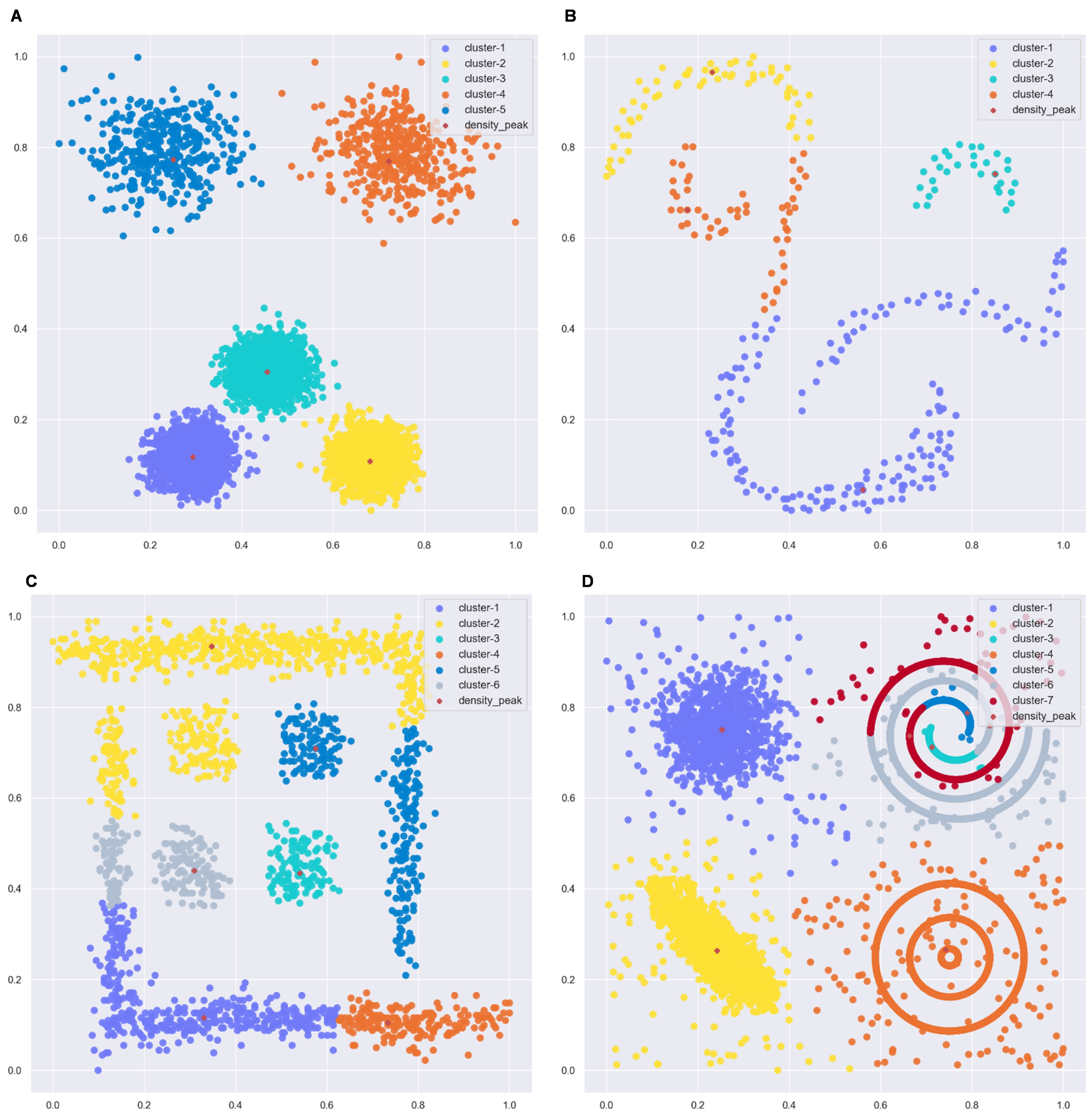}}
		\captionsetup{name={Fig.},labelsep=period,singlelinecheck=off,font={small},justification=raggedright}
		\caption{
			The clustering results of DPeak on 4 synthetic datasets. In all the cases, the value of density peak has been chosen by visual inspection.}
		\vspace{-1.0em}
		\label{fig:nan}	
	\end{figure}
	We have also applied GBC to a significant challenge task, the biomedical image segmentation of immune granular-balls in confocal microscopy on a spatiotemporal biomedical dataset\cite{15,16}. This dataset is particularly relevant for images of immune granular-balls, which exhibit high plasticity and both globular and nonglobular shapes, possibly in contact\cite{17}. In Fig. 8, we show the results of an analysis performed for the biomedical data segmentation of immune granular-balls in confocal microscopy. The ground truth of the biomedical data segmentation of immune granular-balls is displayed in Fig. 8(A). The bar graph Fig. 8(B) shows the Rand Index\cite{18} performance of K-Means, DBSCAN, DPeak, and GBC. This shows that our proposed GBC method achieves the best performance. In Figs. 9(C-F), we depict the clusters with different colors. We can see that K-Means and DPeak split one granular-ball into multiple parts, and disjoint granular-balls are classified into one cluster in Fig. 9(C) and Fig. 9(D). In Fig. 9(E), although DBSCAN can detect the upper cluster in the dotted box, multiple clusters are split into one cluster represented in gray. In contrast, as shown in the dotted box in Fig. 8(F), GBC can detect immune granular-balls well in the biomedical data segmentation, and the detection results are closer to the ground truth than those obtained by other methods.
	
	As our next evaluation, we benchmarked GBC on the Cancer Genome Atlas recognition task using the TCGA dataset\cite{19}, derived from a landmark cancer genomics program. In TCGA, we selected three types of tumor data: aml-mirRNA, kidney-methy, and sarcoma-methy. After linear dimensionality reduction, we obtained clustering results. Combined with survival data, we can generate the survival curves. As shown in Fig. 9, the survival curves based on the clustering results have obvious intervals, and the corresponding log-rank values\cite{20} and Breslow values\cite{21} are significantly less than 0.05, indicating that there are obvious differences between separate clusters and the three types of tumors can be effectively distinguished and identified.
	
	We also applied GBC to a handwriting recognition task using the MNIST dataset\cite{22,23}, a popular benchmark for machine learning algorithms. As shown in Fig. 10(A), this dataset contains ten clusters corresponding to handwritten Arabic numerals. The granular-balls are shown in Fig. 10(B) when the process of granular-ball division converges. The clustering result in Fig. 10(C) shows that our method effectively and adaptively produces ten clusters.
	\begin{figure*}
		\centering
		\subfigbottomskip=-2pt
		\subfigure{
			\includegraphics[width=0.7\textwidth]{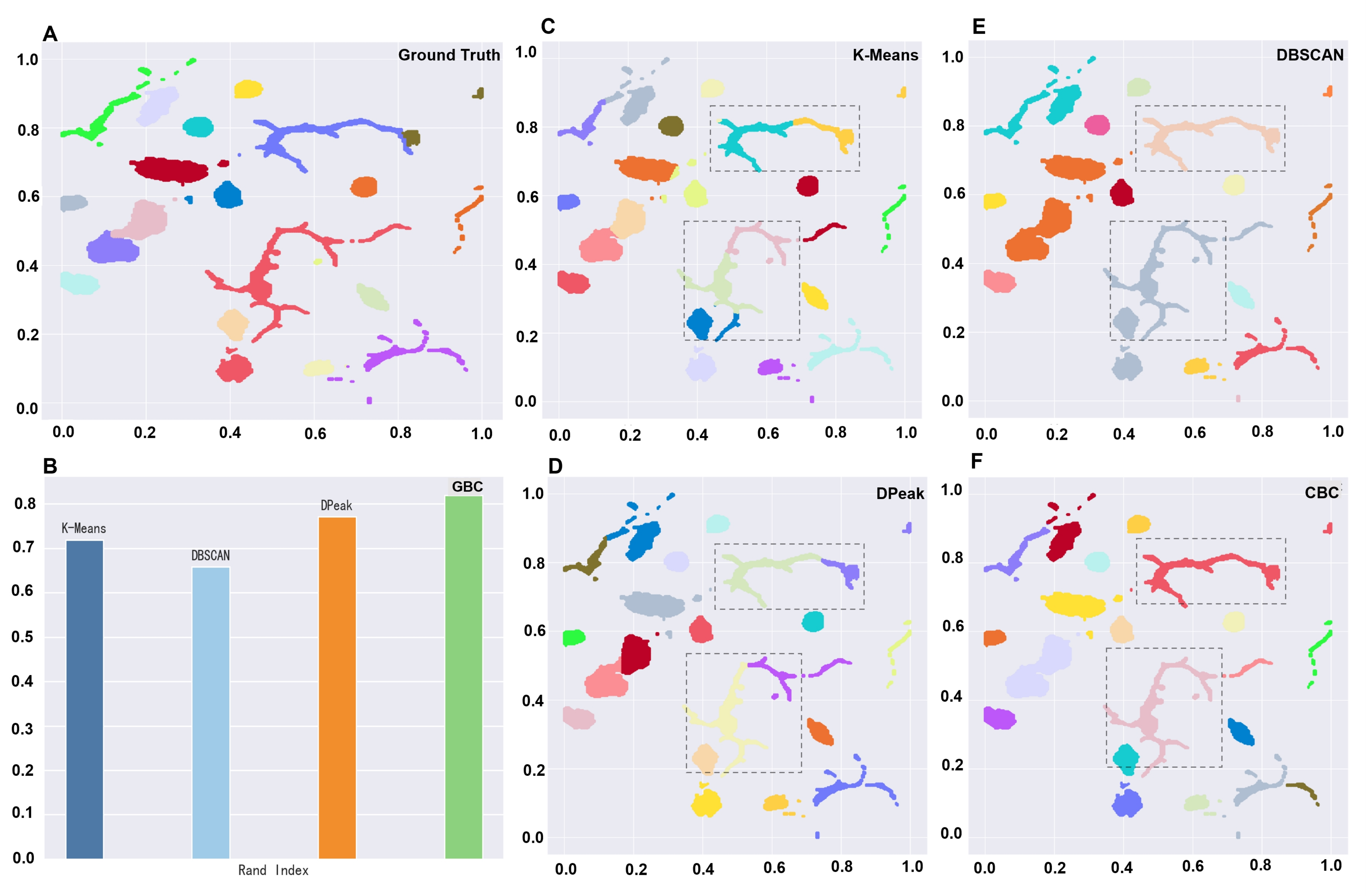}}
		\captionsetup{name={Fig.},labelsep=period,singlelinecheck=off,font={small},justification=raggedright}
		\caption{Cluster analysis of biomedical data segmentation of immune granular-balls in confocal microscopy.  (A) The ground truth of the biomedical data segmentation of immune granular-balls. (B) The performance of K-Means, DBSCAN, DPeak and GBC. (C) The K-Means clustering result. (D) The DPeak clustering result. (E) The DBSCAN clustering result. (F) The GBC clustering result.}
		\vspace{-1.0em}
		\label{fig:nan}	
	\end{figure*}
	\begin{figure*}
		\centering
		\includegraphics[width=0.78\textwidth]{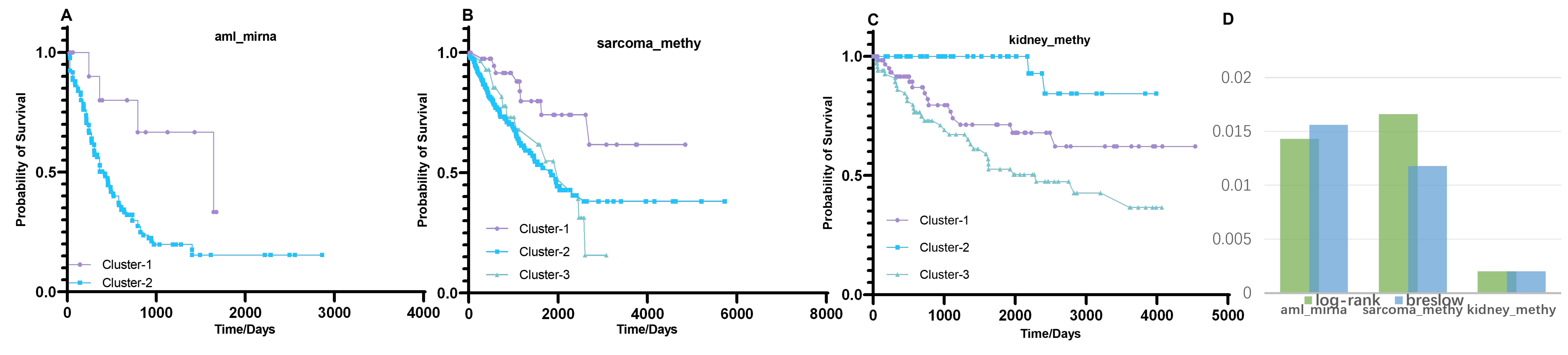}
		\captionsetup{name={Fig.},labelsep=period,singlelinecheck=off,font={small},justification=raggedright}
		\caption{Survival curves and P values of the clustering results of aml\_mirna, kidney\_methy, and sarcoma\_methy data in TCGA.}
		\vspace{-1.0em}
		\label{fig:9}	
	\end{figure*}
	
	\begin{figure*}
		\centering
		\includegraphics[width=1.02\textwidth]{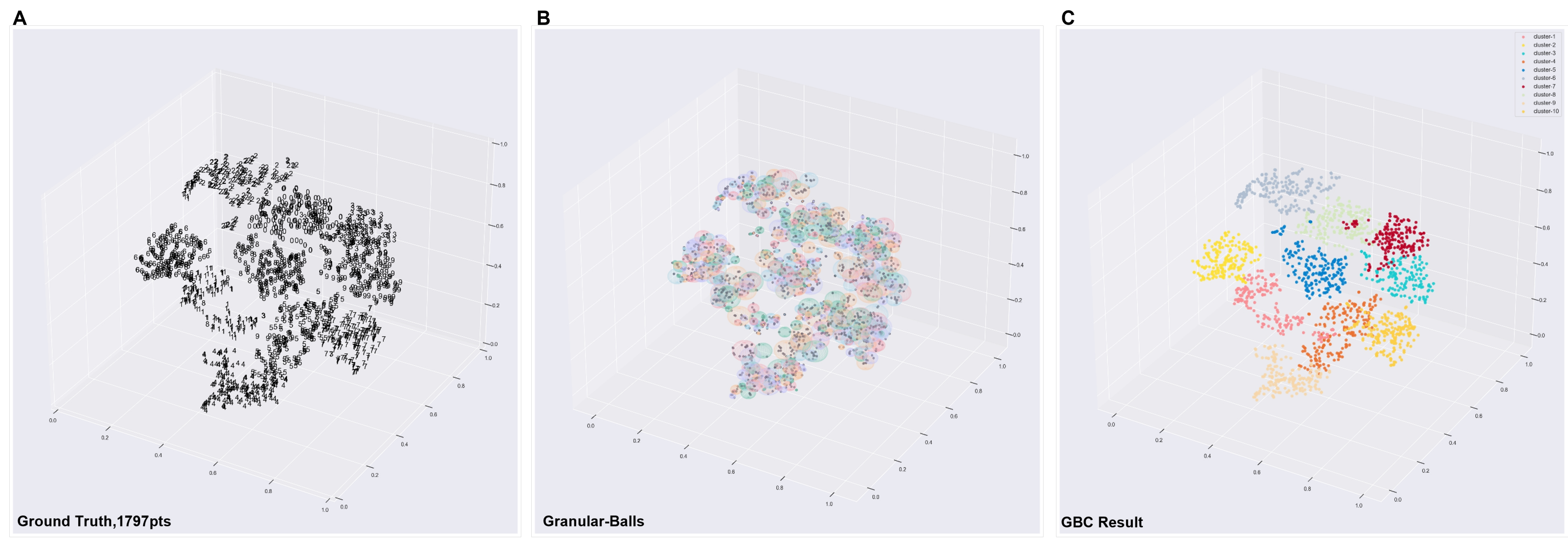}
		\captionsetup{name={Fig.},labelsep=period,singlelinecheck=off,font={small},justification=raggedright}
		\caption{(A-C) Results on MNIST dataset of 1797 with 10 clusters.(A) The ground truth of MNIST dataset is displayed. (B) Granular-ball set (C) Merge the granular-ball sets in  to form final clusters with overlap-rules.}
		\vspace{-1.0em}
		\label{fig:10}	
	\end{figure*}
	
	\begin{table}[H]
		\setlength{\tabcolsep}{2.8mm}{
			\small
			\centering
			\vspace{-1em}
			\setlength{\abovecaptionskip}{0cm}
			\setlength{\belowcaptionskip}{0cm}
			\caption{Comparison of Time (Seconds) Complexity between DP and GBC}
			\begin{tabular}{ccccc}
				\toprule[1pt]
				Data & DP  &DBSCAN &  GBC \\
				\midrule[1pt]
				synthetic data A in figure 5 & 157.8   &20.6 & 3.2         \\
				synthetic data B in figure 5 & 2.1     &0.8 & 0.1           \\
				synthetic data C in figure 5 & 11.2   &2.5 & 8.1          \\
				synthetic data D in figure 5 & 251.3  &25.2 & 2.2            \\
				mnist data                   & 12.3  &3.5 & 0.6            \\
				immune cells data            & 272.9  &50.2 & 5.9            \\
				\bottomrule[1pt]
		\end{tabular}}
		\vspace{-1.0em}
		\label{tab:uci}
	\end{table}
	Most methods which can process datasets with complex shapes are based on density calculations or nearest neighbor calculations (such as DBSCAN and DPeak): from these, only GBC does not need to calculate the distance between each pair of two points. This lends it a speed similar to K-Means, much faster than the other methods used in the experiments, which is demonstrated in Table 1; but unlike GBC, K-Means is unable to process non-convex data or data with a complex distribution, granting GBC an inherent advantage.

	\section{Conclusions and Future Work}
	
	GBC provides an efficient, adaptive method of depicting the world. Depending on the distribution of data, GBC can use granular-balls with appropriate granularity to efficiently and accurately describe the distribution of data. GBC differs from existing non-adaptive methods, in which any number or range of neighbors must be set. This efficient and adaptive method will promote the research and development of adaptive and efficient AI technologies, especially density computing models, and improve the efficiency of a large number of existing clustering methods such as spectral clustering and MST clustering. In addition, GBC is a basic algorithm. We can further improve its performance by introducing some additional technologies, such as kernel technology, non Euclidean distance metric, etc.
	
	\section*{Acknowledgements}
	This work was supported in part by NICE: NRT for Integrated Computational Entomology, US NSF award 1631776, the Natural Science Foundation of Chongqing under Grant Nos. cstc2019jcyj-msxmX0485 and cstc2019jcyj-cxttX0002 and by the National Natural Science Foundation of China under Grant Nos. 62176033 and 61936001.

	\ifCLASSOPTIONcaptionsoff
	\newpage
	\fi
	
	\bibliographystyle{IEEEtran}
	\bibliography{IEEEabrv,sample-base}

\end{document}